\documentclass{article}
\usepackage{spconf_2021,amsmath,graphicx}
\usepackage{indentfirst}
\usepackage{subfigure}
\usepackage{multirow}
\usepackage{amssymb}

\graphicspath{{./images}}

\title{Long-Short Temporal Modeling for Efficient Action Recognition}
\name{Liyu Wu$^{\star}$ \qquad Yuexian Zou$^{\star \dagger}$ \qquad  Can Zhang$^{\star}$\thanks{This paper was partially supported by the IER foundation (No. HT-JD-CXY-201904) and  Shenzhen Municipal Development and Reform Commission (Disciplinary Development Program for Data Science and Intelligent Computing). Special acknowledgements are given to Aoto-PKUSZ Joint Lab for its support. *zouyx@pku.edu.cn }}
\address{$^{\star}$ADSPLAB, School of ECE, Peking University, Shenzhen, China \\
      $^{\dagger}$Peng Cheng Laboratory, Shenzhen, China}
\begin{document}
%
\maketitle
\begin{abstract}
Efficient long-short temporal modeling is key for enhancing the performance of action recognition task. In this paper, we propose a new two-stream action recognition network, termed as MENet, consisting of a Motion Enhancement (ME) module and a Video-level Aggregation (VLA) module to achieve long-short temporal modeling. Specifically, motion representations have been proved effective in capturing short-term and high-frequency action. However, current motion representations are calculated from adjacent frames, which may have poor interpretation and bring useless information (noisy or blank). Thus, for short-term motions, we design an efficient ME module to enhance the short-term motions by mingling the motion saliency among neighboring segments. As for long-term aggregations, VLA is adopted at the top of the appearance branch to integrate the long-term dependencies across all segments. The two components of MENet are complementary in temporal modeling. Extensive experiments are conducted on UCF101 and HMDB51 benchmarks, which verify the effectiveness and efficiency of our proposed MENet.

\end{abstract}
\begin{keywords}
action recognition, video understanding, motion representation, temporal modeling
\end{keywords}
%
\section{Introduction}
\label{sec:intro}
\begin{figure}[t]
\begin{minipage}[b]{.48\linewidth}
  \centering
  \centerline{\includegraphics[width=3.8cm]{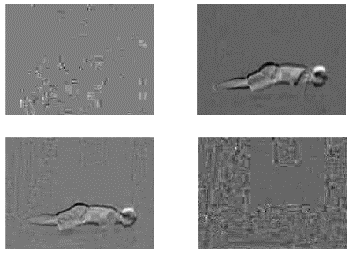}}
  \centerline{(a) PushUps}\medskip
\end{minipage}
\hfill
\begin{minipage}[b]{0.48\linewidth}
  \centering
  \centerline{\includegraphics[width=3.8cm]{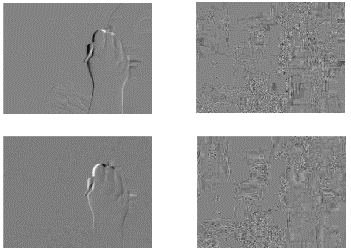}}
  \centerline{(b) ApplyEyeMakeup}\medskip
\end{minipage}
\caption{Under the surveillance of RGBDiff, some of the blocks are noisy and blank, which result in recognizing error.}
\label{fig.1}
\vspace{-1.0em}
\end{figure}

\noindent In the present time, part of the algorithms of action recognition is developing toward a direction of fast and low parameters, where two-stream convolutional networks become one of popular algorithms \cite{Wang2016, Zhang2019, Simonyan2014, Zhou2018_TRN}. Some of the essential features for action recognition are the appearance and motion features. Motion features have been proved significantly effective in action recognition. The networks \cite{Wang2016, Lee2018, Jiang2019, Zhang2019, Simonyan2014} utilize the motion features like optical flow, PA or RGBDiff can achieve the state-of-the-art performance. However, there are a few disadvantages in the existing motion features, (1) Optical flow \cite{10.1007/978-3-540-74936-3_22} suffers from heavy computational overhead and expensive time cost. (2) Due to the redundancy among adjacent frames, motion differences among adjacent frames may have poor interpretation. Fig. \ref{fig.1} explains an example of the redundancy in RGBDiff, in which we could see that some of the blocks in feature maps are noisy and blank in observation.

According to the observation, we propose a method called Motion Enhancement (ME). It is proved valid and efficient in the fields of performance and computation, respectively. ME amplifies the motion features and focuses on the small displacements and entity's contour which are the important factors in action recognition \cite{Sevilla-Lara2019}. Motion representations, however, tend to capture high-frequency motion in short-term temporal relationship. Actions like playing the guitar or field hockey penalty, where the motion movements are almost same and the differences are mainly on the object or background. In this case, it is hard to distinguish by only motion representation without enough appearance information. 

Long-term temporal relationship \cite{Zisserman,He2019, Zolfaghari2018} should be another important issue in video understanding that we concern about. We design a lightweight Video-level Aggregation (VLA) module which applies temporal channel shift technique \cite{Lin_2019_ICCV} to capture long-term temporal relationship in the high-level feature maps of the appearance branch. Additionally, we use some of auxiliary practices to promote the accuracy of the appearance branch in auxiliary. The contributions can be summarized as followed: (1) Propose Motion Enhancement (ME) in order to amplify motion cues and capture short-term temporal relationship. (2) Design Video-level Aggregation (VLA) trying to construct long-term temporal relationship in high semantic feature maps. (3) Extensive experiments demonstrate that our proposed MENet reach the comparable results with the state-of-the-art methods on the several benchmark datasets.

\begin{figure}
    \centering
    \includegraphics[width=0.45\textwidth]{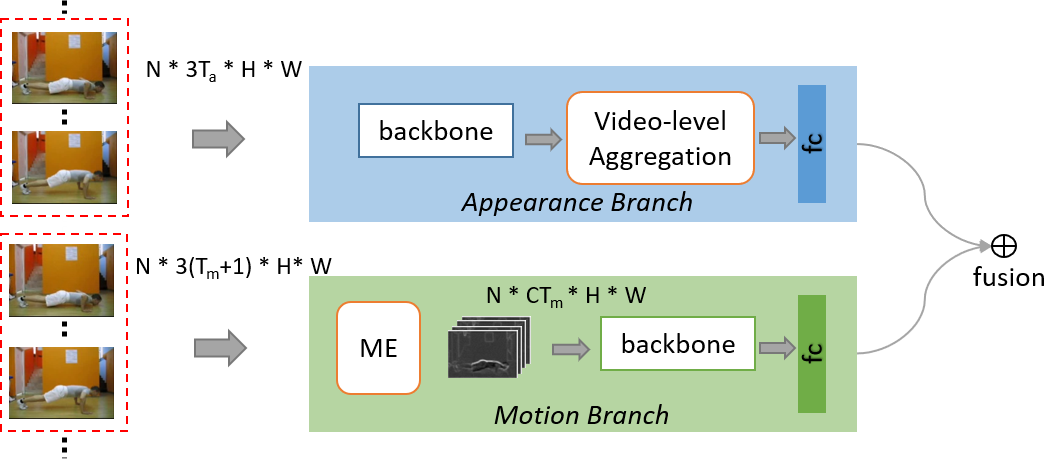}
    \caption{Motion Enhancement Network (MENet).}
    \label{fig.2}
\end{figure}

\vspace{-2.0em}
\section{METHOD}
\label{sec:meth}

\noindent In this section, we introduce our designed two-stream MENet which includes appearance branch and motion branch, showing in Fig.\ref{fig.2}. We describe the detailed concept of the Motion Enhancement (ME) module, which aims to amplify the motion information and model short-term temporal relation. Finally, we introduce a video-level aggregation (VLA) module, which considers the long-term temporal aggregation among all segments in the appearance branch.

\subsection{Short-term Relationship - Motion Enhancement(ME)}
\noindent By introducing very limited computing cost, ME generates efficient motion features to replace the role of optical flow and other motion representations. As mentioned in [6], the small displacements of entity’s motion contour matter most for action recognition. Therefore, we give two adjacent frames to highlight such small motion variations at boundaries. Typically, the brightness consistency constraint of optical flow is defined as follows:
\begin{equation}
\label{eq.1}
    I(x+\Delta{x}, y+\Delta{y}, t+\Delta{t}) \approx I(x, y, t)
\end{equation}
where $I(x,y,t)$ denotes the pixel value of position (x, y) of a frame at time t.  $\Delta{x}$ and $\Delta{y}$ indicate the optimal displacement from time $t$ to  $t+\Delta{t}$ in horizontal and vertical axis separately. In other words, Eq.\ref{eq.1} means the brightness of pixel (x, y) at time t should be approximately equal to the brightness of pixel $(x + \Delta{x}, y + \Delta{y})$ at time $t + \Delta{t}$ under the ideal circumstance. Usually, our goal is to find the optimal displacement $(\Delta{x}^*, \Delta{y}^*)$.

We extend  $I(x,y,t)$ to the feature maps $F(x,y,t)$ in Eq.\ref{eq.2}, then obtain the residual feature space $R(x,y,t)$ between two adjacent feature maps in Eq. \ref{eq.3}.
\begin{equation}
\label{eq.2}
    F(x+\Delta{x}, y+\Delta{y}, t+\Delta{t}) \approx F(x, y, t)
\vspace{-0.5em}
\end{equation}
\begin{equation}
\label{eq.3}
    R(x, y, \Delta{t}) = F(x+\Delta{x}, y+\Delta{y}, t+\Delta{t}) - F(x, y, t)
\end{equation}

Searching the optimal solution of $(\Delta{x}^*, \Delta{y}^*)$ in each pixels is time-consuming without additional constraints such as spatial or temporal smoothness assumptions. Therefore, similar to [4], we restrict our searching place by fixing the directions $\Delta{x},\Delta{y} \in{\{0,\pm1\}},|\Delta{x}|+|\Delta{y}|\leq2$.

Unlike the searching strategy of optical flow, we only care about the motion variations and entity’s contour at certain or neighboring pixel point rather than considering the direction of displacement $(\Delta{x}, \Delta{y})$. There are two conceptions to  support our motion modeling method. First,  low-level feature maps tend to present the general characteristic such as texture, edge and background. Second, with small receptive area, low-level feature have the ability to perceive small area in the input space. Therefore, the differences between low-level feature maps will pay attention on the motion boundaries and the entity’s contour.

Based on the conceptions, we use maxpooling operation on the salient points and amplify the certain points around the neighborhood instead of shift displacements $(\Delta{x}, \Delta{y})$. The new formula could be rewritten as Eq.\ref{eq.5}.
\begin{equation}
\label{eq.4}
    F^{'}_{t+\Delta{t}} = Maxpooling(F(x, y, t+\Delta{t}))
\vspace{-0.5em}
\end{equation}
\begin{equation}
\label{eq.5}
    R(x, y, \Delta{t}) = F^{'}_{t+\Delta{t}} - F(x, y, t)
\end{equation}
 where $F^{'}_{t+\Delta{t}}$ denotes the amplified feature maps. Then each of $F^{'}_{t+\Delta{t}}$ is subtracted by $F_t$. Detailed ME module is demonstrated in Fig. 3. The module helps us generate the salient maps by pixel to pixel between consecutive frames and reflects on the motion representations.
 
 ME module is located at the beginning of motion branch. Formally, we first adopt sparse sample strategy by dividing the video into N segments. For each segment, we randomly sample consecutive $(T_{m}+1)$ frames $F_1, F_2, ..., F_{T_{m}+1} \in R^{3\times{H}\times{W}}$ and pass into {\it Conv2d}. We process the features with maxpooling and denote as $F^{'}_{t+\Delta{t}}$. Pixel-wise value subtraction is computed in $F_t$ and $F^{'}_{t+\Delta{t}}$ in each channel. Finally, we obtain $T_{m}$ frames of motion features and do data permutation to stack those T frames along the channel dimension. ME module, a mapping $R^{(T_{m}+1)\times{3}\times{H}\times{W}} \xrightarrow{} R^{T_{m}C\times{H}\times{W}}$ is built from the appearance to the dynamic motion.

\begin{figure}[t]
    \centering
    \includegraphics[width=5.5cm,height=4.6cm]{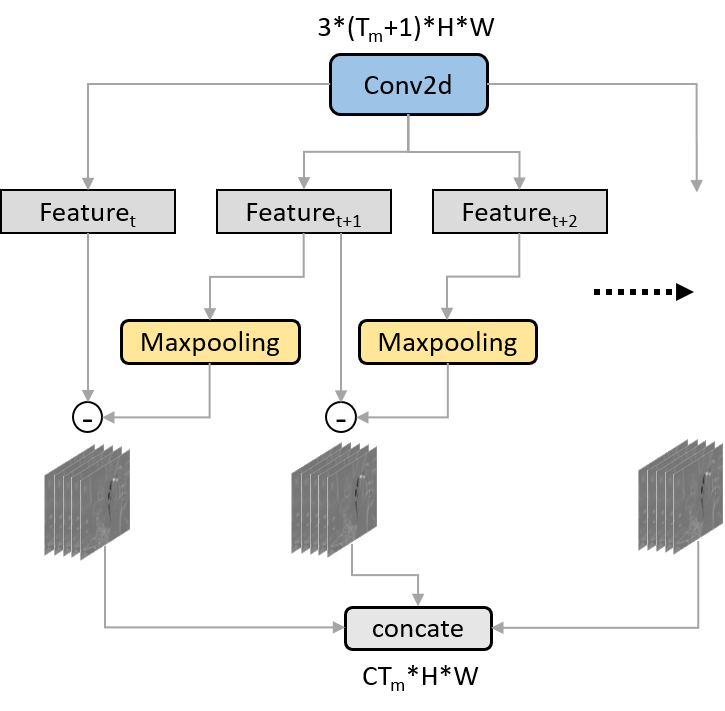}
    \caption{Motion Enhancement (ME) module.}
    \label{fig.3}
\vspace{-1em}
\end{figure}

ME tends to capture the short-term and high-frequency action. Accordingly on the thought, we input two times more segments than the appearance branch, and for the consensus of fusing segments, since we more prefer short-term motion representation, so merely average the score from each segments.
\vspace{-0.5em}
\subsection{Long-term Relationship-Appearance Branch}
\noindent{\bf Video-level aggregation (VLA)}. Considering the long-term temporal relations of all segments, we propose a video-level aggregation (VLA) module to model the high-level feature maps of each segments for better temporal dynamics. VLA module is adopted at the top of the appearance branch as shown in Fig.\ref{fig.2}. It is designed for efficient temporal modeling among feature sequences and easy optimization in an end-to-end manner. Note that \cite{Xie2018} claims that applying temporal convolutions on the top layers of the network is more effective.

\begin{figure}
    \centering
    \includegraphics[width=0.45\textwidth]{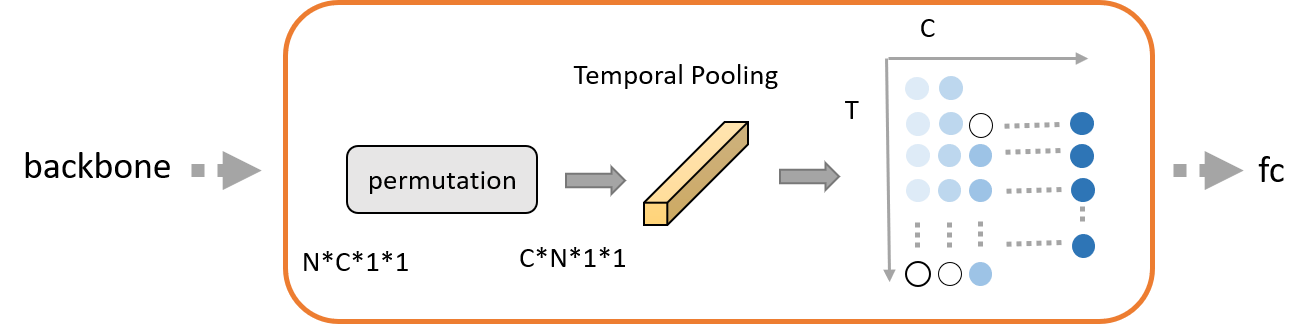}
    \caption{Video-level Aggregation (VLA) module.}
    \label{fig.4}
\vspace{-0.5em}
\end{figure}

As shown in Fig.\ref{fig.4}, assume the original video is divided in to N Segments, and the output activated features after backbone network are ${f_1, f_2, ..., f_N}$, where $f_i \in R^{C\times{H^{'}}\times{W^{'}}}$. We first operate data permutation over the N independent segments into 3D construction by stacking feature maps along temporal dimension for each channel, $f_i^{C\times{H^{'}}\times{W^{'}}} \xrightarrow{} F^{C\times{N}\times{H^{'}}\times{W^{'}}}$. Following by 3D pooling operation, we decouple it into spatial 2D maxpooling and temporal 1D maxpooling. 2D pooling is the same as in ResNet, the extra temporal 1D pooling tries to distill the most important segments over the whole video. The combination of a spatial pooling layer and temporal pooling layer essentially substitutes a 3D pooling operation that extracts activation over spatio-temporal neighborhoods.

We use channel shift \cite{Lin_2019_ICCV} technique to construct long-term range temporal dependency.  Giving the input feature maps after decoupling 3D pooling  $F \in R^{C\times{T}}$, where $T$ denotes the temporal dimension after pooling of $N$, we leverage the shift operation along the temporal dimension. we shift the input $F$ by $-1, 0, +1$ directions and multiply by $\omega_1,\omega_2,\omega_3$ respectively, which sum up to be $Y$. Eq.\ref{fig.6} shows the shift operation, and Eq.\ref{eq.7} interprets the corresponding weights for each shifted channel.
\begin{equation}
\label{eq.6}
F_{i}^{0:c}=X^{0:c}_{i-1},  F_{i}^{c:2c}=X^{c:2c}_{i+1},  F_{i}^{2c:}=X^{2c:}_i
\end{equation}
\begin{equation}
\label{eq.7}
Y = \omega_1 F_{i}^{0:c} + \omega_2F_{i}^{c:2c} + \omega_3 F_{i}^{2c:}
\end{equation}
, where $\omega_1,\omega_2,\omega_3$ could be learned by {\it fc} layer, the superscript c denotes channel index of $F, Y$ is final prediction.

{\bf Network configurations.} In order to give an assistance on modeling long-term temporal relationship, we adopt a few auxiliary practices, including transfer-learning and RGB-Super. As shown in Fig.{fig.6}, motion representation has the property of weak detection, making model more focus around the moving entities. Thus, we initialize the weights of the feature extractor of the appearance branch by the motion branch, compelling the appearance branch to focus around the motion entities. For initializing convenience, we set $3T_a$ equals to $CT_m$, as shown in Fig.\ref{fig.2}, by stacking consecutive frames called RGB-Super \cite{He2019}. Intuitively, input more frames leads to higher accuracy under same computation condition. Our backbone net, ResNet50, can be directly initialized with weights from ImageNet pre-trained model. Weights of Conv1 can be initialized by inflating 2D kernel to 3D size following  I3D \cite{Zisserman}.

\vspace{-1em}
\section{EXPERIMENTS}
\label{sec:experiments}

\subsection{Datasets}

\noindent UCF101 \cite{DBLP:journals/corr/abs-1212-0402} contains 101 human action categories and 13320 labeled video clips collected from the Internet. HMDB51 \cite{hmdb51} is extracted from various sources, including movies and online videos. It contains 6,766 videos with 51 action classes. Kinetics \cite{Zisserman} is the largest well-labeled action recognition dataset containing about 300K videos with 400 human actions. The videos are all collected from the YouTube website.

\subsection{Implement Detail}
\noindent The input modality of MENet is only raw RGB frames. We apply the similar data augmentation strategy as introduced in [1]. All input frames are first resized to 256, then cropped to 224*224 with fixed-corner strategy and scale jittering with horizontal flipping are employed. For the motion branch,we input 16 segments and each segment with 4 consecutive frames, as for the appearance branch, we input 8 segments and each segment with 5 consecutive frames. We train MENet using SGD algorithm, with momentum 0.9 and weight decay 5e-4. The initial learning rate is set as 0.001. The experiments are conducted on one GTX Titan X with 12GB memory. We conduct the experiments on UCF101 and HMDB51 pre-trained on Kinetics400.

\subsection{Experimental Results}

\begin{figure}[htb]
\begin{minipage}[b]{.48\linewidth}
  \centering
    \centerline{\includegraphics[width=3.3cm]{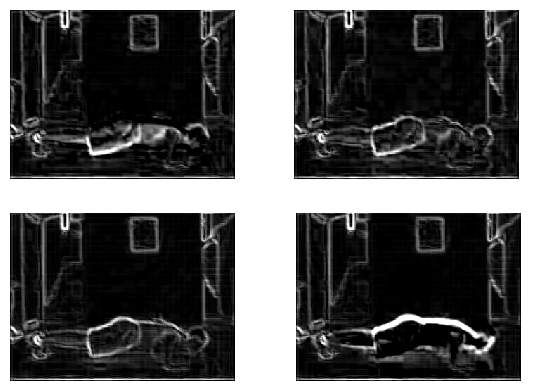}}
  \centerline{(a) PushUps}\medskip
\end{minipage}
\hfill
\begin{minipage}[b]{0.48\linewidth}
  \centering
    \centerline{\includegraphics[width=3.3cm]{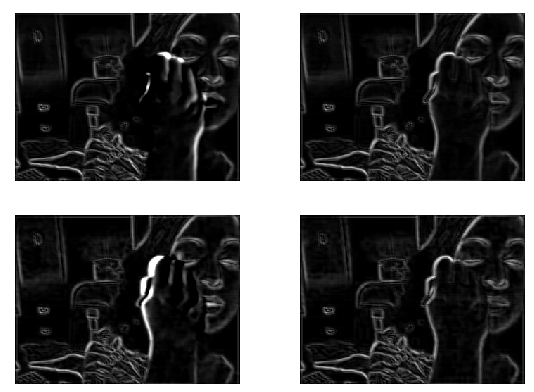}}
  \centerline{(b) ApplyEyeMakeup}\medskip
\end{minipage}

\caption{Motion Enhancement (ME) feature with corresponding frames to Fig.\ref{fig.1} (RGBDiff).}
\label{fig.5}
\end{figure}

\begin{figure}[t]
    \centering
    \subfigure[ME]{
    \includegraphics[width=0.1\textwidth]{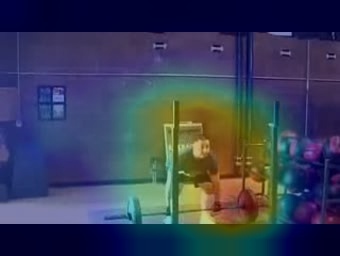}
    \includegraphics[width=0.1\textwidth]{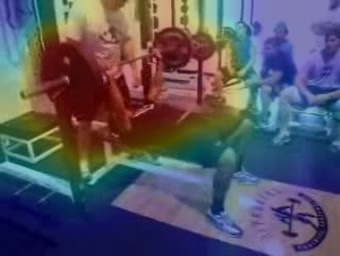}}
    \subfigure[RGB]{
    \includegraphics[width=0.1\textwidth]{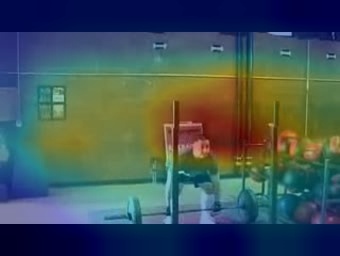}
    \includegraphics[width=0.1\textwidth]{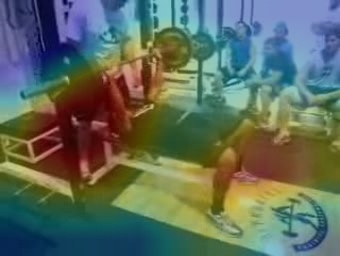}
    }
    \caption{Using grad-cam to observe the dynamic area.}
    \label{fig.6}
\end{figure}

\noindent {\bf Motion branch}. We first visualize the feature maps generated from ME module, as shown in Fig. 5. We could see that in the corresponding blocks of RGBDiff (Fig.\ref{fig.1}) and ME, ME enhances the motion boundaries and entity’s contour and the contour is more smoothness. ME also preserve the background information, in Fig.\ref{fig.5} (b) ApplyEyeMakeup, we could approximately observe face contour, which makes a difference with apply lipstick in the dataset.

Then we compare the performance with other mainstream motion representation. As shown in table 1, for raw RGB input, ME reaches the comparable performance at 87.4\% and 88.3\% for 8 frames and 16 frames respectively. In table 2, while maintaining the accuracy, the speed has been greatly improved and reach 1500fps which is 100 times faster than optical flow \cite{10.1007/978-3-540-74936-3_22}. For deeper analysis, we use grad-cam for visualizing ME. As shown  in Fig. 6, ME focuses more on the action entity while RGB modality focuses more on the background. It proves that we can initialize the appearance branch with this property, which compels the appearance branch to pay more attention on moving area and still concern the background information.

\begin{table}[]
\centering
\caption{Comparison of different modalities on UCF101 split1.}
\begin{tabular}{llll}
\hline
\textbf{Backbone} & \textbf{Modality} & \textbf{Frame} & \textbf{Acc.} \\ \hline
BN-Inception      & RGBDiff \cite{Wang2016}          & 3              & 83.8          \\
ResNet50          & RGBDiff$_{our}$          & 8              & 85.5          \\ 
BN-Incpetion      & Optical Flow \cite{Wang2016}     & 3              & 87.2          \\ 
ResNet50          & RGB               & 8              & 84.9          \\ 
BN-Inception      & PA \cite{Zhang2019}               & 16             & 85.9          \\ \hline
\textbf{ResNet50} & \textbf{ME}       & \textbf{8}     & \textbf{87.4} \\
\textbf{ResNet50} & \textbf{ME}       & \textbf{16}    & \textbf{88.3} \\ \hline
\end{tabular}
\end{table}

\begin{table}[]
\centering
\caption{Comparison of accuracy and efficiency using different motion representations on UCF101 split1.}
\begin{tabular}{lll}
\hline
\textbf{Method}    & \textbf{Acc.} & \textbf{FPS}  \\ \hline
TV-L1 \cite{10.1007/978-3-540-74936-3_22}             & 92.0          & 15            \\
FlowNet \cite{Dosovitskiy_2015_ICCV}           & 86.8          & 52            \\
FlowNet2.0 \cite{2016FlowNet}        & 90.4          & 8             \\
MotionNet \cite{10.1007/978-3-030-20893-6_23}         & 87.5          & 120           \\
Disp. Map(112*112) \cite{Zhao2018} & 88.2          & 190           \\ \hline
\textbf{ME}        & \textbf{88.3} & \textbf{1500} \\ \hline
\end{tabular}
\end{table}

\vspace{-0.5em}

{\bf Appearance branch}. We train multiple variants of our model to show how the performance is improved with proposed configurations. Table 3 shows the network configuration of the appearance branch. While adding VLA and transfer learning strategy, the accuracy is improved over 3\% without extra computation cost. In the experiments, we stack 5 consecutive frames to construct RGB-Super. 

\begin{table}[]
\centering
\caption{Results evaluated on UCF101 split1 validation set with different network configuration }
\begin{tabular}{llll}
\hline
\multicolumn{3}{c}{\textbf{Configuration}}                                                                & \multicolumn{1}{c}{\multirow{2}{*}{\textbf{Acc.}}} \\ \cline{1-3}
\multicolumn{1}{c}{\textbf{RGB-Super}} & \multicolumn{1}{c}{\textbf{VLA}} & \multicolumn{1}{c|}{\textbf{Transfer-learning}} & \multicolumn{1}{c}{}                      \\ \hline
\multicolumn{1}{c}{-}          & \multicolumn{1}{c}{-}    & \multicolumn{1}{c}{-}                   & \multicolumn{1}{c}{84.9}                  \\              
        \multicolumn{1}{c}{\checkmark}                      &           \multicolumn{1}{c}{-}              &           \multicolumn{1}{c}{-}                             & 85.0                                      \\
        \multicolumn{1}{c}{\checkmark}                      &         \multicolumn{1}{c}{\checkmark}                &       \multicolumn{1}{c}{-}                           & 86.9                                      \\
        \multicolumn{1}{c}{\checkmark}                      &         \multicolumn{1}{c}{\checkmark}                &             \multicolumn{1}{c}{\checkmark}                           & 88.2                                      \\ \hline
\end{tabular}
\end{table}


{\bf Two-branch network - MENet}. Combining the motion branch with the appearance branch by using weighted sum, appearances and motions can be used to get the predictions. As shown in table 4, result indicates that the motion branch and the appearance branch may encode complementary information. For fair comparison, only RGB frames as input are reported and the methods are grouped according to their pre-trained datasets. Under the Kinetics \cite{Zisserman} pre-trained strategy, our proposed MENet could achieve the state-of-the-art performance on the two benchmark datasets.

\vspace{-0.5em}
\begin{table}[h]
\centering
\caption{Comparison with state-of-the-art on UCF101 and HMDB51 datasets (over all three splits). }
\begin{tabular}{llll}
\hline
\textbf{Method} & \textbf{Pre-train} & \textbf{UCF101} & \textbf{HMDB51} \\ \hline
Psedo-3D \cite{DBLP:journals/corr/abs-1711-10305}       & Sports-1M             & 88.6            & -               \\ \hline
TSN \cite{Wang2016}            & ImageNet              & 86.4            & 53.7            \\
I3D \cite{Zisserman}           & ImageNet              & 84.5            & 49.8            \\
Disentangling \cite{Zhao2018}  & ImageNet              & 91.8            & -               \\ \hline
ECO \cite{Zolfaghari2018}            & Kinetics              & 94.8            & 72.4            \\
ARTNet \cite{DBLP:journals/corr/abs-1711-09125_AARTNet}         & Kinetics              & 94.3            & 70.9            \\
TSM \cite{Lin_2019_ICCV}            & Kinetics              & 94.5            & 70.7            \\
StNet \cite{He2019}          & Kinetics              & 93.5            & -               \\
CIDC(R2D) \cite{li2020directional}      & Kinetics              & 95.6             &  72.6 \\ \hline
\textbf{MENet}  & \textbf{Kinetics}     & \textbf{95.6}   & \textbf{73.5}   \\ \hline
\end{tabular}
\end{table}

\vspace{-2.0em}
\section{CONCULSION}
\label{sec:conculsion}

\noindent We model long-short temporal relationship of videos by two-branch network. ME module amplifies motion information among adjacent frames. The visualization results verify that ME module significantly enhances the moving entity’s contour that is critical for action recognition task. It is worth noting that ME module not only improves accuracy, but also achieves 100 times faster motion modeling speed than traditional optical flow methods. On the other hand, VLA module uses temporal shift operation to model long-term aggregations with zero parameters. The two modules of the MENet are complementary in temporal modeling. Experiments on two benchmarks demonstrate the proposed MENet reaches current state-of-the-art methods, and verify our design intuition.

\vfill\pagebreak



\bibliographystyle{IEEEbib}
\label{sec:refs}

\end{document}